\documentclass[11pt]{elsarticle}

\usepackage{palatino,setspace,graphicx,rotating,mathrsfs,bm,listings,caption}
\usepackage{subfigure,latexsym,amsmath,amssymb,natbib, rotating,
  ocg-p}
\usepackage[pdftex,colorlinks=true,citecolor=blue,raiselinks=false]{hyperref}

\makeatletter
\long\def\@makecaptiohen#1#2{%
  \vskip\abovecaptionskip
  \sbox\@tempboxa{#1: #2}%
  \ifdim \wd\@tempboxa >\hsize
    #1: #2\par
   
  \else
    \global \@minipagefalse
    \hb@xt@\hsize{\box\@tempboxa\hfil}%
  \fi
  \vskip\belowcaptionskip}

\makeatother

\begin{document}

\author{Wendy K. Tam\fnref{wkt}}  
  \ead{wendy.k.tam@vanderbilt.edu}

  \fntext[wkt]{Wendy K. Tam
  is Professor and Stevenson Chair in the Departments of Political Science, Computer
  Science, Biomedical Informatics, and the Law School at Vanderbilt University, and
  an affiliate of the National Center for Supercomputing
  Applications at the University of Illinois at Urbana-Champaign}
  
\title{{\LARGE {\bf The Amplifying Mirror}}  \\ {\large Locating and Steering the Partisan Direction inside a Large Language Model}}
            
\begin{abstract}
\begin{singlespace}
Large language models are rapidly replacing search engines as the primary interface between people and information.  Unlike search engines, which retrieve existing content, LLMs generate novel text shaped by internal representations learned during training.  Here we show that partisan political identity is encoded in the model's activation space, and that this direction directly shapes generation.  Using 190,491 tweets from sitting members of U.S. Congress as labeled training data, we train linear probes on the hidden states of the Llama 3.1 8B Instruct model.  We identify a single geometric axis at layer 18 that separates Republican from Democratic text with an AUC of 0.945 and a Cohen's $d$ of 1.94, and use sparse autoencoders to decompose that axis into interpretable partisan features.  Causally intervening along this axis---ablating or amplifying the partisan component mid-generation---produces systematic shifts in the model’s output.  We witness stance reversals, register shifting, and structured fabrications of authority.  Our results demonstrate that partisan bias in language models is not a vague emergent property but a learned geometric feature that can be precisely located and steered.  Partisan bias is not a bug to be patched, but a structural property of how these models encode information about their users.  As LLMs displace search engines as the interface to knowledge, understanding that product design (and its consequences) will be essential for navigating the legal, social, and political transitions from an information ecosystem that is curated to one that is generated. \\ \\

\noindent
\end{singlespace}
\end{abstract}

\begin{titlepage}
  \maketitle
  \pagestyle{empty}
  \setcounter{page}{0}

\end{titlepage}

\pagestyle{headings}


\clearpage
\newpage

\section{Introduction}

Since the release of ChatGPT in November 2022, large language models (LLMs) have rapidly become a major interface between people and information.  In a massive paradigm shift, the information they provide is generated rather than retrieved.  That distinction matters.  A search engine {\em retrieves} human-authored content that can be traced to identifiable sources and evaluated as such.  An LLM, by contrast, {\em generates} text with no identifiable human author, and through internal mechanisms that remain, at best, only partially understood.

LLMs do not always produce the same answer to the same query.  Beyond the stochastic variation introduced by sampling, LLM outputs can vary systematically with cues about {\em who} is asking~\citep{Sharmaetal:24}, reflecting and sometimes amplifying stereotypes that the model has absorbed about different kinds of users~\citep{Koteketal:23, Chenetal:26, Argyleetal:23, Santurkaretal:23}.  The question, then, is not only {\em what} these models say, but {\em why} they say it and {\em how} they arrive at what they say.  If a language model builds a representation of its user and conditions its output on that user model, individual users are receiving different responses to identical prompts, not because the information landscape is fragmented, but because the model, itself, is acting as a mirror that reflects the user's inferred identity back at them.

We examine {\em why} the same prompt can produce different outputs for different users.  Using tools from mechanistic interpretability, which seeks to reverse-engineer the internal computations of neural networks, we also investigate {\em how} those outputs are produced~\citep{Cunninghametal:24, AlainBengio:18, Parketal:24}.  We show that partisan political identity is represented in the Llama 3.1 8B Instruct model as a specific, locatable direction in its activation space.  Using tweets from sitting members of the US Congress, we train linear probes on the model's hidden states to isolate a single direction that separates Republican from Democratic text~\citep{MarksTegmark:24}.  We then use sparse autoencoders to decompose that direction into interpretable features.  Causally intervening along this identified direction during model generation, either by ablating or by amplifying it toward either party, produces systematic and predictable changes in output~\citep{Turneretal:23, Lietal:23}.  Our interventions yield clear stance reversals on politically valenced prompts, broad shifts in rhetorical register, and, in some cases, the fabrication of named political figures and organizational outlets that are consistent with the steered direction.  We thus demonstrate that partisan representation is not merely a diffuse emergent property, but a geometric feature of the model’s representational space that can be identified, decomposed, and causally manipulated.

\section{Detecting and Locating the Partisan Signal}

We begin with the basic question of whether an LLM encodes partisan information at all.  More specifically, can it infer a user's partisan proclivity from text alone?  To help answer this question, we employ the PoliticalTweets dataset that is downloadable from the Hugging Face website (\href{https://huggingface.co/datasets/Jacobvs/PoliticalTweets}{https://huggingface.co/datasets/Jacobvs/PoliticalTweets}).  The data set contains 190,491 tweets posted by sitting members of the U.S. Congress on X/Twitter between September 2016 and February 2023.  In addition to the text of the tweet, the data set includes the posting date, tweet ID, Twitter handle, and the member's party affiliation.  Using only the text of a single tweet at a time, we prompted Llama 3.1 8B Instruct to identify the author's partisan identity.  For this task, the model correctly classified the tweeter's party 64.4\% of the time, indicating that a reliable and statistically significant partisan signal resides somewhere inside the model, though it reveals no information about where that signal resides inside the network or how the model uses it.\footnote{Note that the accuracy is larger than implied by the number since many tweets carry no partisan signal.  For instance, Rand Paul (R--KY) has a tradition of releasing his Festivus Report and tweeting ``Happy Festivus'' every year in December.  See, e.g., https://x.com/RandPaul/status/1474101158209859593}

The Llama 3.1 8B Instruct model has 32 transformer layers, each with a 4,096-dimensional hidden state.  To locate the partisan signal, we collect the hidden-state activations at every one of the model's transformer layers for all 190,491 tweets, which amounts to approximately 6.6 TB of raw activation data and an additional 50 GB of extracted last-token vectors for downstream analysis.  At each layer, we then train a logistic regression probe to predict the tweeter's party from the residual-stream activation values~\citep{Belinkov:22}.  Our goal is to identify whether the partisan signal crystallizes into a linearly separable representation at a particular layer depth, which would give us a principled intervention site.  If it does, we then ask   whether that information is concentrated in a single direction or is distributed across many directions.  Because transformer features are typically packed into activation space through superposition~\citep{Elhageetal:22}, the partisan signal will need to be disentangled from neighboring features.

\begin{figure}[htbp]
  \centering
  \includegraphics[width=5.5in]{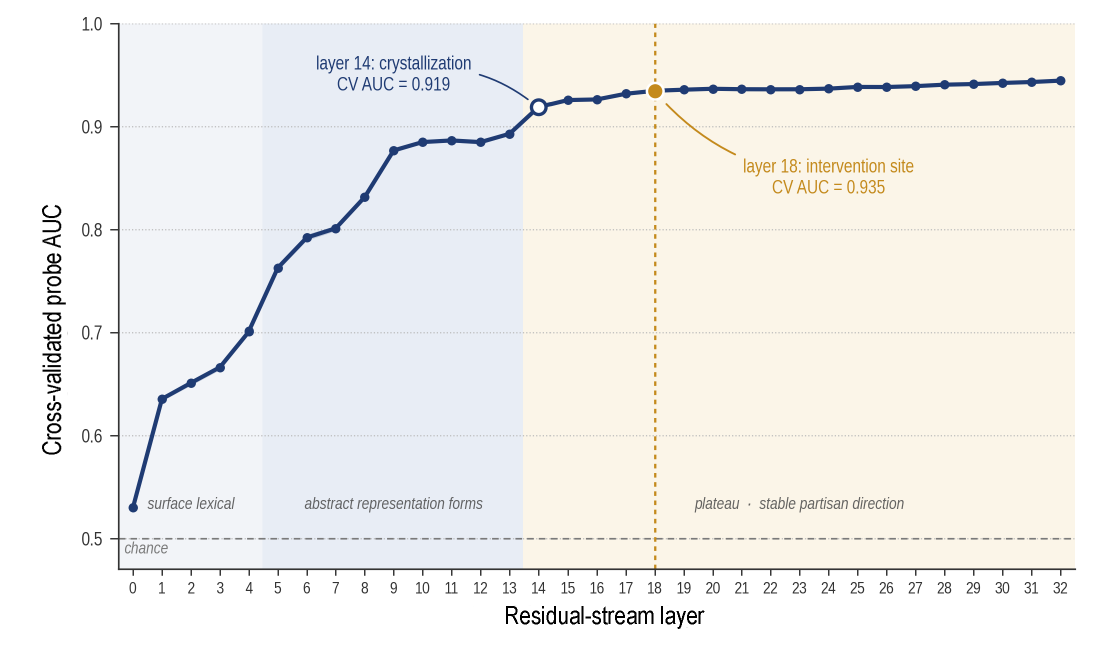}
  \caption{Per-layer cross-validated probe AUC across all residual-stream layers of the Llama 3.1 8B Instruct model}
  \label{fig:auc_by_layer}
\end{figure}

Figure~\ref{fig:auc_by_layer} shows our results.  The earliest layers (0--4) perform near chance (cross-validated AUC 0.53--0.70), suggesting that they capture little beyond surface lexical cues.  Through layers 5--13, an abstract political representation forms (AUC 0.67--0.89).  By layer 14, the probe AUC crosses 0.92 and the representation crystallizes and stabilizes.  After this point, the cross-validated probe AUC rises with only minor fluctuations from 0.92 to 0.945 over the remaining layers.\footnote{The probe AUC (0.945) exceeds the model's zero-shot classification accuracy (64.4\%) by quite a bit.  The two metrics measure different things.  Zero-shot accuracy assesses whether the model produces the correct partisan label for a tweet.  The probe, on the other hand, asks whether partisan information exists in the model's {\em internal representation}, regardless of whether this representation surfaces during generation.  The gap between the two is both interesting and informative because it indicates that the model encodes substantially more partisan signal than it explicitly reports.  That is, it uses the partisan signal implicitly by its framing, tone, and source selection, yet not reliably routing that into the correct classification token.  As explained earlier, many tweets carry no partisan signal, which depresses classification accuracy but does not penalize the probe, which simply assigns ambiguous cases a score near the decision boundary.  AUC rewards this uncertainty while accuracy does not.}  Partisanship, once encoded, appears to be preserved through the rest of the network.  From layer 14 onward, probe directions also show strong stability across cross-validation folds, with cosine similarity approaching or exceeding 0.90.

\begin{figure}[htbp]
  \centering
  \includegraphics[width=5.5in]{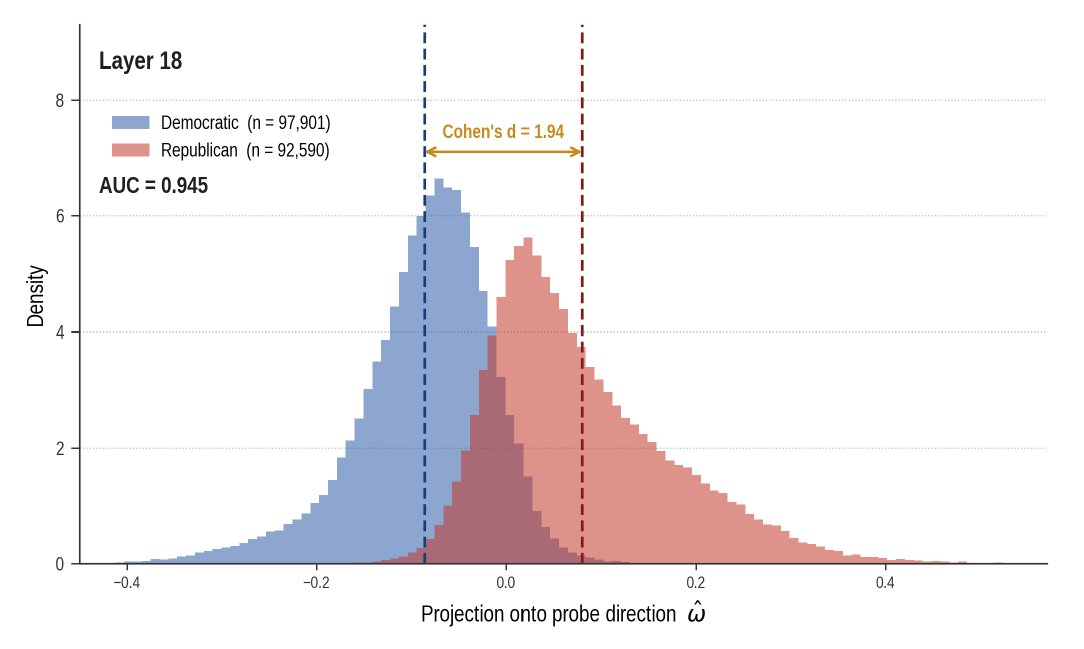}
  \caption{Distribution of layer 18 activations projected onto the learned probe direction, $\hat\omega$, for tweets from Republican (red) and Democratic (blue) Members of Congress.}
  \label{fig:layer18_hist}
\end{figure}

A simple mean-difference direction (the normalized vector $\hat\mu_R - \hat\mu_D$) at layer 18 achieves an AUC of only 0.715, which is substantially below our trained probe's AUC of 0.945 at the same layer.  This gap suggests that the partisan representation is multi-dimensional.  That is, a single axis defined by the class means does not, by itself, capture the representation~\citep{Gurnee:23}.  Yet, once the relevant geometry is learned from data, the signal becomes strongly linearly separable as shown in Figure~\ref{fig:layer18_hist}.  When activations are projected onto the probe's learned weight vector, $\hat{\omega}$, the Republican and Democratic distributions at layer 18 separate with a projection AUC of 0.945 and a Cohen's $d$ of 1.94, nearly two standard deviations. (The cross-validated probe AUC, which is a more conservative estimate, is 0.935.)

We chose layer 18 as our intervention site.  Although the deepest layer yields the highest AUC, an intervention there would leave no downstream computation through which the perturbation could propagate, because any change at that layer would feed directly into the final token representation.  Intervening at layer 14, by contrast, would risk disturbing a representation that has only begun to stabilize.  Layer 18 provides a practical middle ground.  By this point in the model, the partisan direction is both stable and cleanly separable, and there remain a number of downstream layers through which the intervention can shape generation.  We next ask whether our identified direction plays a causal role in shaping the partisan bent of the model's outputs.

\section{Steering the Model}

If $\hat\omega$ is the direction along which the model represents partisanship, then modifying activations along $\hat\omega$ during generation should produce predictable changes in output.  We test two interventions on the layer 18 hidden state.  {\em Ablation}, $x' = x - (x \cdot \hat\omega) \, \hat\omega$, projects the activation onto the hyperplane orthogonal to $\hat\omega$, thereby removing the partisan component.  {\em Amplification}, $x' = x + \alpha \, \hat\omega$, where $\alpha$ is scaled in proportion to the activation's existing projection onto $\hat\omega$~\citep{Rimskyetal:24}, pushes the activation farther along the same direction.

\begin{figure}[htbp]
  \centering
 \includegraphics[width=\textwidth]{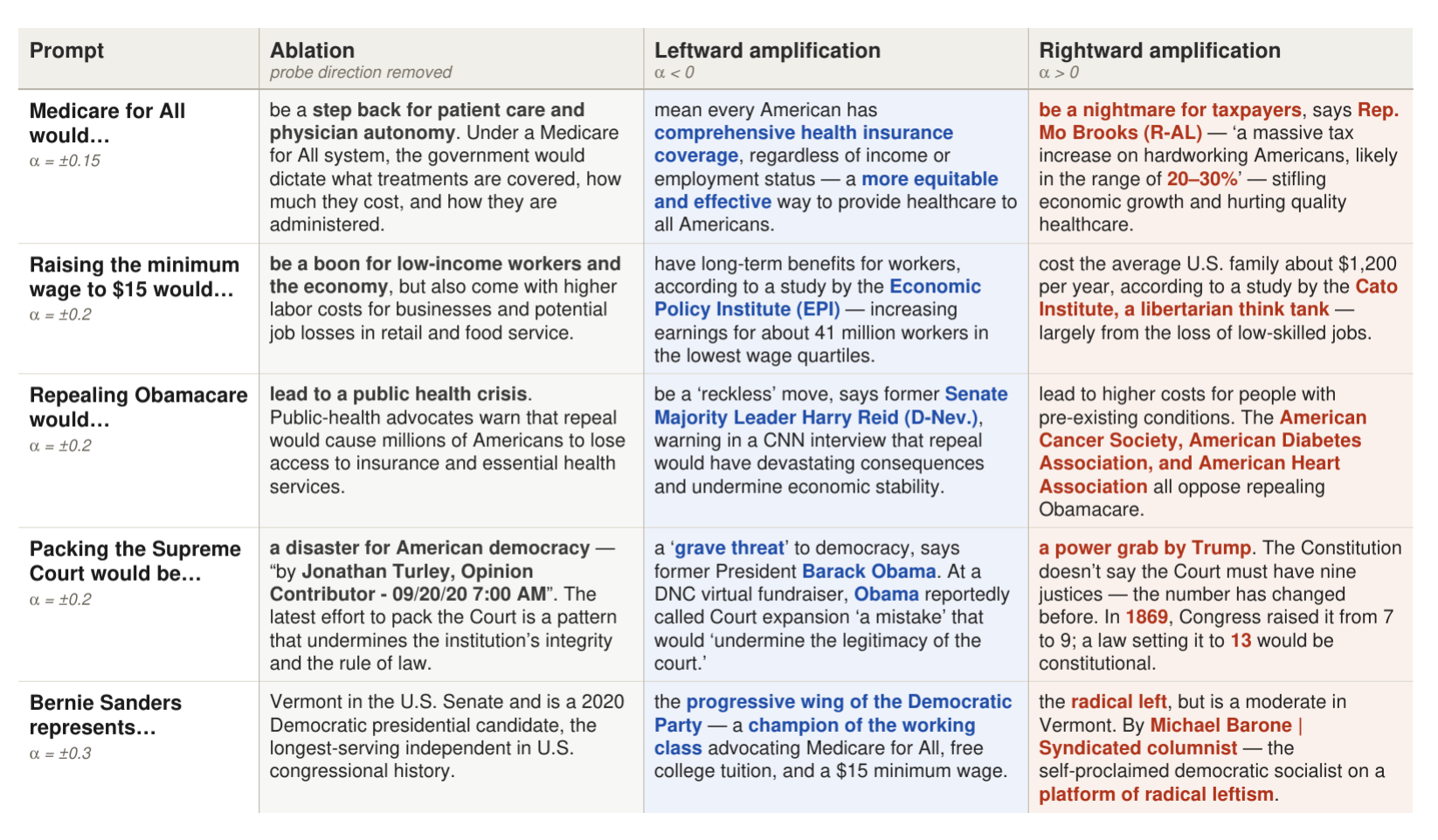}
 \caption{Example outputs under ablation and bidirectional amplification}
 \label{fig:exemplars}
\end{figure}

We test these interventions on 150 prompts spanning a range of contested issues (e.g., health care, taxes, affirmative action, and international affairs), where the two major parties have staked out strongly opposing positions and where each prompt admits a plausible completion in either partisan direction.  All of our interventions use fractional amplification strengths with $\alpha_{\rm_{frac}} \in [0.15, 0.3]$, corresponding to perturbations of 15–30\% of the hidden-state norm.  Figure~\ref{fig:exemplars} displays five of our prompts with the LLM completions under ablation, leftward amplification, and rightward amplification.  Across our set of prompts, both interventions produce systematic and predictable changes in output, which fell into three recurring patterns.  

The first notable pattern is {\em stance inversion} on prompts where there is a possible binary policy conclusion.  For the prompt, ``Medicare for All would \dots'' (row 1, $\alpha_{\rm frac} = \pm 0.15$), leftward amplification completes the prompt with universal ``comprehensive health insurance coverage'' that is ``more equitable and effective,'' while rightward amplification produces ``a nightmare for taxpayers.''  For the prompt, ``Raising the minimum wage to \$15 would \dots'' (row 2, $\alpha_{\rm frac} = \pm 0.2$), leftward amplification cites an Economic Policy Institute study finding benefits for ``about 41 million workers in the lowest wage quartiles,'' whereas rightward amplification cites a Cato Institute study estimating a \$1,200 annual cost per family.  The prompt is identical.  The model is the same.  The opposite conclusions are the direct result of a causal manipulation of the activation values at layer 18.

Second, we observe {\em register shifting}, where the same underlying content is reframed in the argumentative style, vocabulary, and emotional valence associated with the two political camps.  For the prompt ``Repealing Obamacare would \ldots'' (row 3, $\alpha_{\rm frac}= \pm 0.2$ ), both directions yield opposition to repeal, but the leftward completion is delivered as a named politician quotation (a ``\,`reckless' move, says former Senate Majority Leader Harry Reid (D--Nev.)''), while the rightward completion grounds the same position in a coalition of medical professional associations (the American Cancer Society, American Diabetes Association, and American Heart Association).
For the prompt, ``Bernie Sanders represents …'' (row 5, $\alpha_{\rm frac} = \pm 0.3$), ablation returns a bland biographical description (``Vermont in the U.S. Senate and is a 2020 Democratic presidential candidate'').  Leftward amplification describes him as ``the progressive wing of the Democratic Party\dots \, a champion of the working class,'' and rightward amplification reframes him as ``the radical left.''

Third, and with greater societal consequences, in our opinion, we observe {\em named-entity fabrication} where, rather than merely becoming more opinionated, the model generates specific political figures, advocacy organizations, or media outlets aligned with the steered direction, often with invented attributed quotations.  In row 1, we see that rightward amplification at $\alpha_{\rm frac}=0.15$ (which was the gentlest intervention we tested) of ``Medicare for All would\dots'' produced a quotation attributed to Rep.~Mo Brooks (R--AL) with a frabricated and specific numeric claim that the program would result in a ``massive tax increase\dots likely in the range of 20-30\%.''  In row 5, rightward amplification reframes its completion as a syndicated column by the conservative commentator Michael Barone.  In row 4 (``Packing the Supreme Court would be\dots''), leftward amplification attributes a specific ``grave threat'' phrasing to former President Obama at a DNC virtual fundraiser.  Ablation of that prompt produces a fabricated byline (``Jonathan Turley, Opinion Contributor, 09/20/20 7:00 AM'').  The attributed individuals are real, and their broad political alignments are correct, but the specific statements are not sourced from any utterance of theirs that we could identify.  This makes partisan steering a double-edged intervention since the probe direction carries enough semantic content to consistently select real, camp-aligned speakers, while also producing confident misattribution, which is a more consequential form of distortion than either stance inversion or register shifting.  Across our broader set of 150 prompts, leftward fabrications include attributed statements from Drs.~Anthony Fauci, Rochelle Walensky, Peter Hotez, Leana Wen, and Senator Debbie Stabenow (D–MI).  Rightward fabrications include attributed statements from Dr. Marc Siegel and Dr. Paul Offit.

Lastly, the ablation results in Figure~\ref{fig:exemplars} are instructive but uneven.  For the Bernie Sanders prompt, removing the partisan component produces recognizably neutral output.  Ablation there yields a factual biographical description with no discernible partisan framing.  But other ablations retain a detectable lean.  ``Medicare for All would\dots'' completes as ``a step back for patient care and physician autonomy,'' which is a recognizably conservative framing.  ``Raising the minimum wage should be\dots'' produces  balanced output, while ablation for ``Repealing Obamacare would be\dots'' and ``Packing the Supreme Court is\dots'' retains left-leaning framings.  There are at least three, not mutually exclusive, explanations for this residual lean.  First, the model itself may be leftward leaning in that either the alignment training optimized for responses its human (more leftward-inclined) raters preferred or progressive framing is simply more prevalent in the training corpus.\footnote{This residual leftward lean is consistent with a growing body of evidence that LLMs exhibit a systematic liberal orientation~\citep{Rozado:24, Motokietal:24, Santurkaretal:23, Fengetal:23}.}  In either case, the resulting bias would then be distributed throughout the model's weights rather than concentrated on a single axis.  If so ablating in a particular direction would be unable to remove what was never localized there.  Second, a single linear probe may not capture the full geometry of partisan representation.  Partisanship in the model may occupy a subspace that has more than one dimension, and our intervention removes only the component aligned with the dominant axis we identified.  Our results are consistent with the possibility that our probe has identified a dominant partisan direction while other partisan remnants may remain.  Third, our ablation zeroes out the projection onto $\hat{\omega}$ at layer 18, but the partisan signal may re-emerge in downstream layers through redundant encoding or nonlinear interactions that the linear probe did not capture.  That ablation works as expected on some prompts but not others may also suggest that some topics occupy a space that is closer to our identified axis than others.  Ablation along $\hat{\omega}$ should thus be understood as an attenuation of partisan content, but not a guarantee of neutrality.  Left and right steering seem to align with our expectations, but further study is needed to explore how better to ablate the partisan signal.

\subsection{Steering ``Non-partisan'' prompts}

\begin{figure}[htbp]
  \centering
 \includegraphics[width=\textwidth]{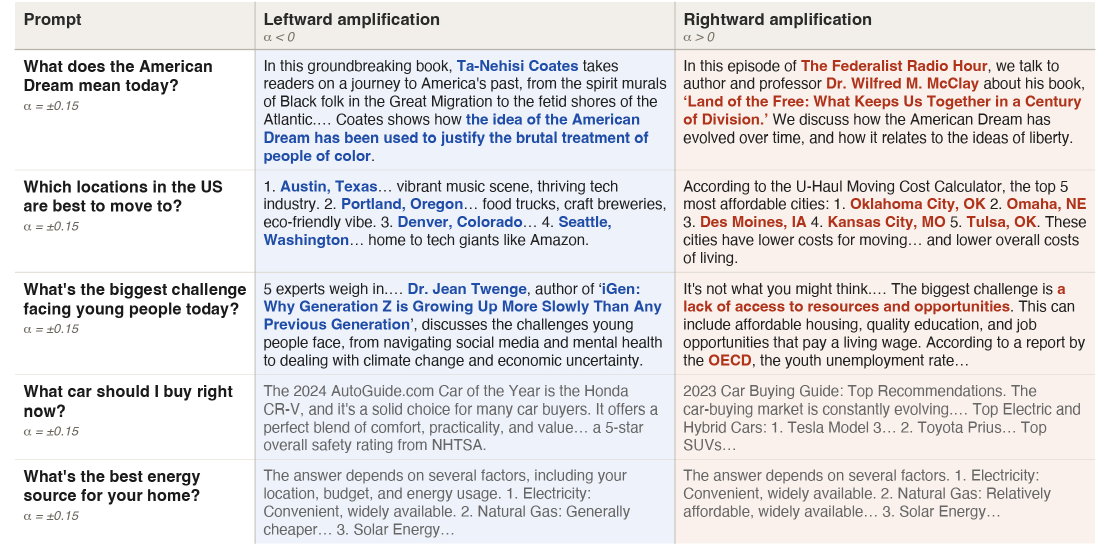}
 \caption{Bidirectional amplification for ``non-partisan'' prompts.}
 \label{fig:nonpartisan}
\end{figure}

It is also instructive to note that the partisan direction does not activate only on overtly political prompts.  Figure~\ref{fig:nonpartisan} presents steering results for five prompts that contain no explicit political content.  The top three prompts produce markedly different outputs under leftward and rightward amplification.  Under leftward amplification, the American Dream prompt generates a response centered on Ta-Nehisi Coates and how the ideal ``has been used to justify the brutal treatment of people of color'', while rightward amplification frames the answer around an episode of {\em The Federalist Radio Hour} and a book about liberty and national unity.  So, even when the prompt queries abstract cultural concepts, and does not hint at policies, the partisan direction can route the model into entirely different intellectual traditions, with different cited authorities and different normative assumptions.  The relocation prompt shows a similar pattern where leftward amplification recommends relocation to Austin, Portland, Denver, and Seattle (cities generally regarded as progressive) while rightward amplification recommends Oklahoma City, Omaha, Des Moines, Kansas City, and Tulsa (emphasizing affordability and cost of living).  

The remaining two prompts requesting recommendations for cars and energy sources are unaffected by steering in either direction.  The selective partisan activation seems to fire on prompts that are culturally loaded.  This implies that the model's partisan representation is not a blunt bias applied uniformly to all output, but a structured encoding that activates selectively when the prompt touches territory where partisanship has been infused in the training data.

Although a single direction recovered by our linear probe is sufficient to steer the model’s partisan output, it tells us little about the internal vocabulary through which the model represents partisanship.  Does the direction, $\hat\omega$, capture a coarse summary of a single monolithic ``political feature,'' or does it instead aggregate many narrower features?  A trained probe cannot distinguish between these possibilities.  It establishes only that a separating hyperplane exists.

\section{Decomposing the Partisan Direction}

To inspect the internal structure of our identified direction and decompose the partisan direction into an interpretable vocabulary, we train sparse autoencoders (SAEs) on the model's hidden-state activations~\citep{Cunninghametal:24, Brickenetal:23}.  An SAE learns an overcomplete dictionary in which each activation is reconstructed from a small number of active features, allowing us to recover a sparse, discrete code for each tweet and then to probe that code directly.  Our SAEs expand the 4,096-dimensional residual stream eight-fold, to 32,768 features.  To assess how much of the partisan signal is captured by this dictionary, we compare probe performance in the SAE latent space with the probe performance on the raw activations.  The gap between the SAE latent probe AUC and the raw-activation probe AUC indicates how much of the signal is lost in the dictionary.  If there is no gap, that means that the dictionary captures the partisan representation in full.

Our initial training runs used standard L1 sparsity regularization and had low reconstruction error, but the learned codes were not sparse.  Out of our 32,768 features, more than 27,000 fired on the average activation.  This implies that these vanilla autoencoders learned a dense representation that reconstructed activations well but were not sparse.  By contrast, the TopK architecture~\citep{Gaoetal:24}, which constrains each activation to use at most $k$ features, achieved sparsity with acceptable reconstruction quality.  The results reported here are from the TopK SAEs with $k = 64$.

We trained SAEs in three configurations: per-layer, PCA-reduced, and pooled across middle layers. The PCA-reduced variants lagged the per-layer configuration by approximately 2 AUC points at every layer. The pooled middle-layer approach was competitive with the per-layer SAEs at intermediate depths, but it discarded the layer-specific structure needed for targeted intervention at a single layer. For our purposes, the per-layer configuration was therefore the most useful.

\begin{figure}[htbp]
  \centering
  \includegraphics[width=5.5in]{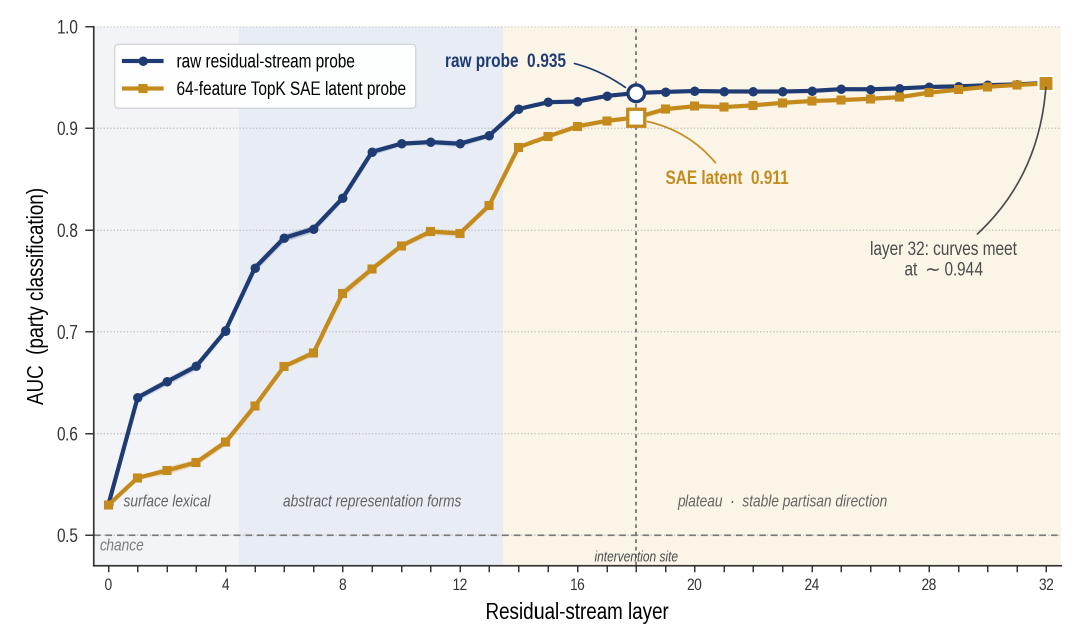}
  \caption{Per-layer comparison of the raw-activation probe AUC (navy) and the 64-feature TopK SAE latent AUC (gold) across all residual-stream layers.}
  \label{fig:sae_vs_probe}
\end{figure}

Figure~\ref{fig:sae_vs_probe} shows the per-layer comparison between the raw-activation probe AUC and the 64-feature TopK SAE latent AUC across all residual-stream layers.  The SAE's AUC climbs from 0.53 at layer 0 to 0.944 at the final layer, tracking the raw probe's AUC but lagging behind it through the middle layers.  At layer 18, our intervention site, the raw probe achieves a cross-validated AUC of 0.935, whereas the SAE code reaches 0.911, a gap of just over two AUC points.  By the final layer, however, the two curves converge at approximately 0.94.  This pattern suggests that a TopK SAE constrained to 64 active features is not sufficient to fully express the partisan representation at the middle layers, but is sufficient to capture it by the final layer.  Partisanship thus appears to begin as a distributed representation, spread across more directions than a sparse code of this size can efficiently capture, but becomes progressively consolidated over the course of the network into a smaller set of salient features that our TopK SAE is able to recover.

Inspecting the individual SAE features at layer 18 confirms this picture of a partially crystallized vocabulary.  Among the ten features with the largest logistic regression coefficients, five are recognizable ideological content features that activate almost exclusively on tweets from a single party.  One Republican-leaning feature (feature 9036, $\beta = 2.88$) fires on anti-administration rhetoric centered on the IRS, immigration enforcement, censorship, and energy policy.  Its top thirty activating tweets are all in the Republican direction.  One Democratic-leaning feature (feature 19268, $\beta=-2.99$) activates on voting rights and democracy framing.  It is likewise 30/30 Democratic.  Three other high-weight features (27699, 27872, and 12677) split 30/30, 29/30, and 28/30 along similarly coherent thematic lines.  These features read less like low-level lexical statistics than like human-recognizable political talking points.  The remaining features in the top ten are less interpretable.  Feature 7612, for example, activates on high-conflict rhetoric from both parties, splitting 16/14 across its top thirty examples.  It appears in the ranking because of its activation variance rather than any consistent directional signal.  As we saw from Figure~\ref{fig:sae_vs_probe}, at layer 18, a sparse code with only 64 active features captures most, but not all, of the partisan signal.  So what it recovers is a hybrid of crystallized political content and residual noise from neighboring dimensions that the SAEs cannot yet disentangle.

\section{Discussion}

Partisan political bias in the Llama 3.1 8B Instruct model is not a diffuse emergent property.  Rather, it is a locatable and steerable geometric feature of the model.  We have shown that a single direction in a single layer's activation space predicts partisan identity with an AUC of 0.945.  Pointedly, intervention along that direction reliably flips stances, rewrites rhetorical register, and conjures specific political figures and institutions appropriate to the steered direction.  When the model infers partisan tilt in its interlocutor, it returns output consistent with that partisan orientation.

The societal stakes of this new technology come into focus against a longer history of communication infrastructure.  Preceding communication technologies---the printing press, broadcast radio, cable television, the Internet, and social media---reshaped public discourse in ways its designers did not anticipate, and each did so by altering how information circulates in society~\citep{ProcacciniTam:27}.  Cable television shattered the shared national narrative~\citep{Prior:07}.  Social media algorithms created echo chambers by amplifying content for user engagement~\citep{Sunstein:17}.  LLMs mark a qualitative break from that lineage.  They do not merely select or amplify existing human content; they generate new content, conditioned on the model's inferred picture of the user.  Our steering experiments provided concrete evidence of not only this tendency, but also how the nature of this model design manifests in the output generation.  

When our interventions caused structured fabrications, the result was not a hallucination in the usual sense---a random failure of recall that produces confident nonsense~\citep{Jietal:23}.  It was a {\em structured} fabrication; the model selected real, camp-appropriate authority figures and attributed plausible but invented statements to them.  The partisan direction evidently carries enough semantic content to reliably match speakers to positions, which means the same geometric feature that encodes ideology also encodes the social structure of who says what and for which side.  The outputs read as authoritative sourcing for positions the user was already expected to hold.  This ``failure mode'' is particularly concerning because it produces precisely the kind of source-attributed claims that are the most difficult for readers to verify and most likely to be believed~\citep{PennycookRand:21}.  Where prior communication technologies filtered or amplified existing human speech, an LLM can manufacture speech that never occurred, attribute it to a real person, and {\em calibrate} it to the user's inferred political identity.

Prior work has found that political perspective can be linearly decoded from individual attention heads in a language model~\citep{Kimetal:25}, which is a useful observation, but one that leaves open where in the network the signal consolidates, what kind of training signal best captures partisan language, and what happens when the identified direction is used to intervene on generation.  We address all of these critical gaps.  First, we probe the full residual stream at each layer, rather than just the outputs of individual attention heads, thereby capturing the cumulative contribution of both attention and feed-forward subnetworks.  Second, we use partisan tweets rather than roll-call vote scores as ground truth.  A vote score places a legislator on an ideology axis, but it carries no information on how partisans talk, which is precisely what an LLM generates.  Third, we demonstrate not only how the model can be steered, but also characterize the linguistic artifacts produced by our intervention, including the fabrication of authoritative sources.  That phenomenon has direct implications for misinformation research and has not, to our knowledge, previously been characterized or explored.

Our study has several limitations, and each points toward a direction for future work.  First, we examine a single model.  We chose the Llama 3.1 8B Instruct model because it is an open-weight model, which permits full access to internal activations, and its 8-billion-parameter scale also makes the full pipeline tractable on a single GPU.  All experiments were run on one NVIDIA H200 GPU.  Even on that hardware, the amount of computation was nontrivial.  Extracting residual-stream activations for 190,491 tweets across all layers produced 6.6 TB of stored activations, and the full pipeline, including 48 GB of extracted last-token vectors and over 100 GB of trained SAE checkpoints, consumed roughly 7 TB on disk.  Training 70 TopK sparse autoencoders with 32,768 features for 50 epochs each, together with 5-fold cross-validated probes at every layer, required 50+ GPU hours end-to-end, and settling on that pipeline consumed 1,000+ GPU hours over the development period.  Larger or newer models may distribute the same concept across multiple layers or encode it with a different geometry.  They may resist intervention in ways that an 8B instruction-tuned model did not, and later alignment training may patch the particular behaviors we document.  These possibilities bound what we can claim here, but they also define the comparative agenda that our method makes tractable.  Llama 3.1 8B is small enough to be fully instrumented while still exhibiting the political reasoning and user modeling of production-scale systems.  If partisan structure is already this legible at this scale, the question for larger models is not whether the structure exists but how it is distributed.  Every open-weight release opens the possibility for a new observation.

Second, our probe defines the partisanship concept operationally.  The direction, $\hat\omega$, is whatever separates the tweets from Republican and Democratic Members of Congress, which may not map perfectly onto other notions of partisanship or political leaning.  However, the close correspondence between our identified direction and its causal effects suggests that our operationalization captures a genuine feature of the model's computation, and not merely a statistical artifact of the training data.  Nevertheless, our construct can be refined to explore neighboring constructs such as ideology rather than party or how other constructs might interact with a partisan signal.

Third, locating a mechanism is not the same as knowing what to do about it.  The pattern of partisan output we identify is troubling for an already polarized society, and is likely a consequence of upstream design choices rather than an {\em intent} to polarize {\em per se}.  Alignment training optimizes for user satisfaction.  Pre-training corpora are saturated with politically charged discourse.  The architecture builds a representation of its interlocutor to devise how best to craft a response.  Each choice is individually reasonable and defensible.  Together, however, they produce a system that calibrates responses to a user's inferred identity, which reinforces existing beliefs rather than advancing the shared understanding that enables democratic self-governance.  That the bias is structural rather than intentional makes it harder, not easier, to address since there is no single decision to reverse and no obvious place in the pipeline where the problem originates.  The search for a remedy, then, is a research question rather than a policy switch.  The steering apparatus that demonstrated the problem can evaluate potential corrections, including deployment-time ablation and targeted fine-tuning, against a measured baseline, and comparisons of base and aligned model can locate where in the pipeline the partisan calibration enters.  Though we do not offer a solution here, we do offer an instrument for testing potential solutions.

A language model infers a picture of its interlocutor and generates a response calibrated to that picture.  In that broad sense, this behavior is not unfamiliar.  Human communicators also adjust their speech to a mental model of their listener.  In conversation, we respond after assessing the other's knowledge, their beliefs, their biases, and what kinds of arguments are most likely to be well-received.  We draw on a shared world in which both speaker and listener live; and listeners are aware that speakers are accommodating them.  An LLM has no such grounding.  Its user model is built from statistical regularities in a training corpus, and the mechanism by which those regularities determine the next token is not visible by default.  Neither the user nor the model is aware of how that representation shapes generation.  Our results show that this mechanism is nevertheless discoverable.  A user's inferred partisan identity can be traced to particular model activations and causally manipulated.  But this discovery required the full apparatus of mechanistic interpretability applied to an open-weight model.  For the hundreds of millions of people who use these systems daily, the mechanism is invisible.

The lack of transparency is the core concern.  LLMs are rapidly displacing search engines as the primary interface between people and information, yet they operate on a fundamentally different principle.  A search engine retrieves content that already exists; a language model generates text conditioned on an inferred user.  LLMs are not neutral information sources.  They are not a ``better Google.''   Using these systems well therefore requires an understanding and awareness of {\em how} they generate their output.  This information is also essential to any serious policy response.  What these models do is a function of human design choices.  Our aim is to make those choices more visible by showing where and how LLMs bend the light.

\section*{Acknowledgements}
\label{sec:ack}

\begin{singlespace}
\begin{small}
\noindent
This work used the NCSA Delta and DeltaAI Supercomputer at the University of Illinois at Urbana-Champaign through allocation CIS260312 from the Advanced Cyberinfrastructure Coordination Ecosystem: Services \& Support (ACCESS) program, which is supported by U.S. National Science Foundation grants \#2138259, \#2138286, \#2138307, \#2137603, and \#2138296.

\end{small}
\end{singlespace}

\clearpage
\newpage

\vspace{-7mm}
\begin{singlespace}
\bibliographystyle{apsr}
\bibliography{mirror}
\end{singlespace}

\end{document}